\documentclass[10pt,twocolumn,letterpaper]{article}

\usepackage[pagenumbers]{wacv} 

\usepackage{graphicx}
\usepackage{amsmath}
\usepackage{amssymb}
\usepackage{multirow} 
\usepackage{booktabs}
\usepackage{pifont}
\newcommand{\xmark}{\ding{53}}%

\usepackage[pagebackref,breaklinks,colorlinks]{hyperref}

\usepackage[capitalize]{cleveref}
\crefname{section}{Sec.}{Secs.}
\Crefname{section}{Section}{Sections}
\Crefname{table}{Table}{Tables}
\crefname{table}{Tab.}{Tabs.}

\newcommand{\encoder}{\mathcal{E}}

\def\wacvPaperID{1400} 
\def\confName{WACV}
\def\confYear{2025}

\begin{document}

\title{DiffSTR: Controlled Diffusion Models for Scene Text Removal}

\author{Sanhita Pathak\\
BSTTM, IIT Delhi\\
{\tt\small Sanhita.Pathak@dbst.iitd.ac.in}
\and
Vinay Kaushik\\
Dept. of CSE, IIIT Sonepat\\
{\tt\small vkaushik@iiitsonepat.ac.in}
\and
Brejesh Lall\\
Dept. of EE, IIT Delhi\\
{\tt\small brejesh@ee.iitd.ac.in}
}
\maketitle

\begin{abstract}

To prevent unauthorized use of text in images, Scene Text Removal (STR) has become a crucial task. It focuses on automatically removing text and replacing it with a natural, text-less background while preserving significant details such as texture, color, and contrast. Despite its importance in privacy protection, STR faces several challenges, including boundary artifacts, inconsistent texture and color, and preserving correct shadows. Most STR approaches estimate a text region mask to train a model, solving for image translation or inpainting to generate a text-free image. Thus, the quality of the generated image depends on the accuracy of the inpainting mask and the generator's capability. In this work, we leverage the superior capabilities of diffusion models in generating high-quality, consistent images to address the STR problem. We introduce a ControlNet diffusion model, treating STR as an inpainting task. To enhance the model's robustness, we develop a mask pretraining pipeline to condition our diffusion model. This involves training a masked autoencoder (MAE) using a combination of box masks and coarse stroke masks, and fine-tuning it using masks derived from our novel segmentation-based mask refinement framework. This framework iteratively refines an initial mask and segments it using the SLIC and Hierarchical Feature Selection (HFS) algorithms to produce an accurate final text mask. This improves mask prediction and utilizes rich textural information in natural scene images to provide accurate inpainting masks. Experiments on the SCUT-EnsText and SCUT-Syn datasets demonstrate that our method significantly outperforms existing state-of-the-art techniques.

\end{abstract}

\vspace{-0.5cm}
\section{Introduction}
\label{sec:intro}
Natural scene images encompass a considerable amount of sensitive and confidential data, including names, addresses, and mobile numbers in the images. The advanced OCR techniques \cite{nguyen2021dictionary} can automatically and effortlessly extract such information that can lead to the unauthorised usage of sensitive text information. In order to prevent such leakage of confidential data, the problem of Scene Text Removal (STR) \cite{nakamura2017scene} has been introduced. STR is tasked with the automatic removal of text regions and replacing them with a consistent background of natural image.
Recently, STR has attracted many applications in the computer vision community because of its valuable applications in privacy protection \cite{inai2014selective}, visual translation \cite{singh2021textocr}, information reconstruction \cite{liu2020textual}, and image editing \cite{yang2020swaptext}.

Although STR provides significant importance in data protection, it faces several challenges \cite{wang2021pert}. The textual content, when removed from the scene, leaves the boundary remnants as artifacts \cite{CTRNet}. In cases where the boundary artefacts are not observed, the text removal region doesn't get the consistent texture and color similar to the text's neighbourhood region\cite{du2023modeling}.

Several approaches have tried to solve the STR task as an image translation or generation problem with a one-stage, two-stage or iterative process. This paradigm incorporates using GAN architectures to achieve realism in image generation\cite{tursun2019mtrnet,tursun2020mtrnet++}. However, optimal training of GANs is quite challenging. Moreover, GANs are unable to generate images of optimal structural consistency and high fidelity \cite{rombach2022high}. 


Recently, Diffusion models\cite{ho2020denoising,rombach2022high,song2020denoising} have gained prominence due to their training stability, high-quality and diverse sample generation, comprehensive mode coverage, straightforward objective function, better interpretability and control, and robust conditional generation capabilities\cite{ho2020denoising}. These models also perform significantly better than GANs in terms of texture quality, image realism and fidelity.

\begin{figure}
    \centering
    \includegraphics[width=0.9\linewidth]{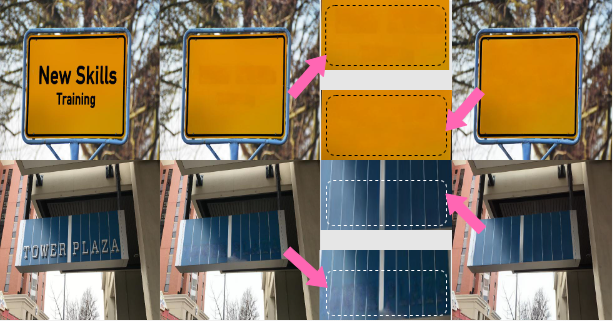}
    \caption{From left-to-right: Original scene image, Text erased image by conditional step, closer view of text background after erasure, DiffSTR result}
    \label{fig:enter-label}
    \vspace{-0.5cm}
\end{figure}

The applications of diffusion in methods as text-to-image models \cite{nichol2021glide,ramesh2021zero,rombach2022high} leveraging multimodal conditional inputs has made utilisation of complex inputs possible. Several methods, such as Imagic\cite{kawar2023imagic} and Null-Inversion\cite{mokady2023null}, have investigated the potential of utilising pretrained diffusion models for image modification while maintaining the integrity of specific areas of interest. 

While Diffusion models have demonstrated impressive outcomes in general image editing\cite{rombach2022high,cao2022learning}, applying them to the specific task of scene text removal is not a simple process. The Diffusion models demonstrate fidelity and good generation capability but they still struggle with structure and texture consistency and are not as efficient without context.  
Introducing this context in Diffusion models for scene text removal is a challenging step, since the missing or incorrect context may further lead to texture inconsistency in generated images. To solve this, we utilise ControlNet as a conditioning input to the diffusion model for guiding the structure and texture generated in the inpainting process\cite{yu2018generative}. 
With the goal of imposing a soft conditioning, we introduce masked image modelling to generate a coarse inpainted image that conditions the Diffusion model. This can be facilitated by finetuning a Masked Autoencoders(MAE) on the data to be predicted(textless image). 



To the best of our knowledge, DiffSTR is the first work to leverage power of conditional diffusion models for scene text removal. The contributions of our work are as follows:
\begin{itemize}
    \setlength{\itemsep}{0pt} 
    \setlength{\parskip}{0pt} 
    \setlength{\parsep}{0pt}  
    \item We leverage diffusion models to solve the problem of scene text removal, achieving high-fidelity and texture-consistent scene generation. 
    \item We propose a novel approach to solving the scene text removal (STR) problem as an inpainting task using a conditioned latent diffusion model. This is achieved by incorporating coarse conditioning within the diffusion model, guiding the text removal process and ensuring texture consistency in the generated images.
    \item We develop a masked pretraining pipeline for generating coarse text-less image, conditioning the diffusion model by training a masked autoencoder. 
    \item We introduce a segmentation-based mask refinement framework to improve the accuracy of inpainting masks for text removal.
    \item Extensive experiments demonstrate that our method outperforms existing state-of-the-art approaches in scene text removal.
    
\end{itemize}

\section{Related Works}
\textbf{Text Removal Methods}
Scene text removal methods \cite{huang2014robust,epshtein2010detecting} initially retrieved text using either a colour histogram or threshold-based algorithms to compute similarity between pixels and replace the background subsequently.

With the introduction of deep learning techniques, many methods have utilised it for inpainting and text detection to process scene text images. The Scene Text Eraser \cite{nakamura2017scene} deals with the removal of text from scenes using Convolutional Neural Networks (CNN).
\cite{zdenek2020erasing} and \cite{conrad2021two} employ a 2 stage methodology which utilises a scene text detector with predictive modelling for segmentation mask and an inpainting network,  trained independently. 
\cite{tursun2019mtrnet} presents the concept of MTRNet, which incorporates a text area mask as an extra input in the network and \cite{tursun2020mtrnet++} expands upon the capabilities of MTRNet.
\cite{zhang2019ensnet} proposes an end-to-end architecture that utilises a GAN with a UNet encoder-decoder generator using multiple loss functions. \cite{keserwani2021text} presents a comprehensive network adversarial architecture conditioned on the text areas. PERT\cite{wang2021pert} employs a systematic approach to effectively eliminate text content and \cite{liu2020erasenet} presents the GAN-based EraseNet utilising hierarchical erasing structure with a segmentation component. \cite{bian2022scene} presents an end-to-end architecture comprising a network for detecting text strokes and a network for generating text removal. FETNET\cite{LYU2023109531} incorporated text segmentation maps at feature level using the proposed FET mechanism. SAEN\cite{du2023modeling} presented a network that focuses on modeling text stroke masks that provide more accurate locations for erasing text. 
ViTEraser\cite{peng2024viteraser} implicitly integrates text localization and inpainting with a pretraining method, termed SegMIM, which focuses on the text box segmentation and masked image modeling tasks, respectively.

\textbf{Image Inpainting}
Image inpainting is a technique that requires coherent pixels to fill in the mask area while maintaining color accuracy, which makes it a complex task. The use of low-level image attributes limits classical techniques like patchmatch \cite{barnes2009patchmatch} and partial differential equation\cite{bertalmio2000image} to produce appropriate outcomes. Generational Adversarial Network (GAN)\cite{goodfellow2014generative} inpainting methods \cite{cao2022learning} use adaptive convolutions \cite{liu2018image}, attention mechanisms \cite{ko2023continuously}, and high-resolution generalisation in frequency learning \cite{chu2023rethinking}. Co-Mod \cite{zhao2021large} uses adversarial learning for plausible generation to solve inpainting.
Techniques that prioritise bigger reconstruction penalties \cite{cao2022learning} yield more consistent results but struggle with big masks, resulting in fuzzy defects. Diffusion models \cite{rombach2022high} have advanced and produced outstanding inpainting results in recent years. These models align distributions rather than reconstructing pixel space, like GAN-based methods. Despite improved generation, Diffusion-based methods suffer from context-instability issues.

\textbf{Diffusion models}
Diffusion models have shown promising results when compared with GANs in image generation. Several works\cite{rombach2022high} show impressive results for image-to-image translation. Further research \cite{avrahami2022blended} has shown that its also possible to control the generated image with optimal conditioning. In brief, using the pre-trained language-vision model such as CLIP \cite{patashnik2021styleclip},  \cite{kim2022diffusionclip} shows the good results in image manipulating task using CLIP as a loss function. Diffusion models have been used in text generation and editing and shown promising results when compared to GAN based counterparts. This motivated us to utilise Diffusion Models for Scene-text-removal task.

\textbf{Masked Image-Modeling}
Masked Image-Modeling (MIM) is a highly studied field in the realm of self-supervised learning. Conventional MIM approaches \cite{chen2024context,he2022masked} learn to generate masked patches based on visible patches. The training targets for visible patches include pixel values \cite{he2022masked}, HOG features \cite{wei2022masked}, and high-level semantic features \cite{wei2022mvp}. Although the main goal of MIM is to learn representations, its potential for generating images is considerable. \cite{cao2022learning} utilises MAE features and attention scores to enhance the convolutional inpainting model's ability to effectively handle long-distance dependencies.
The proposed method employs Masked autoencoder (MAE) taking the motivation of its application in other tasks\cite{wang2024contextstablevisualconsistentimageinpainting}, beforehand to improve the stability of the context in diffusion models.

\begin{figure*}
    \centering
    \includegraphics[width=0.9\linewidth]{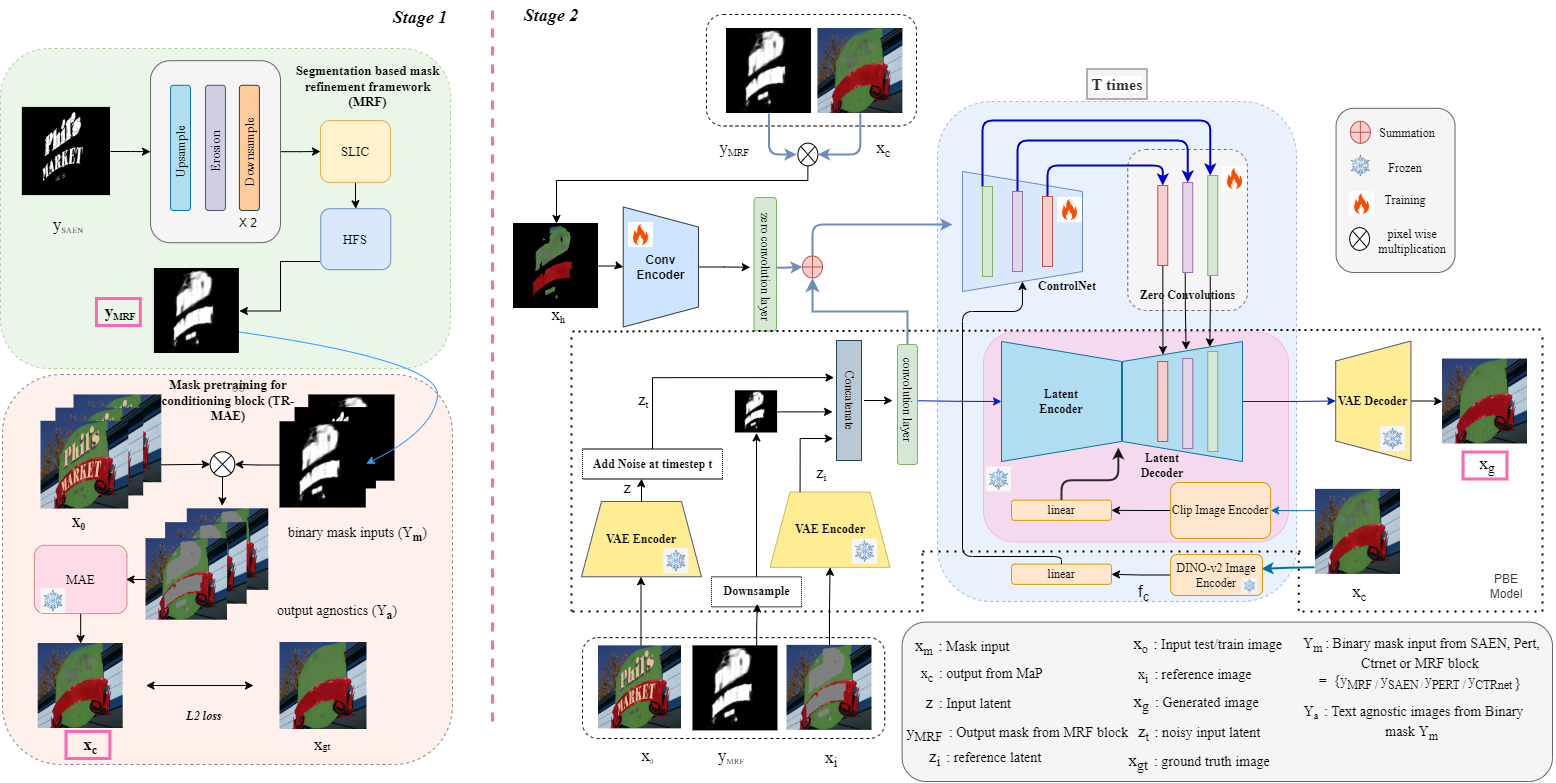}
    \caption{Architecture diagram of DiffSTR. The pipeline consists of two stages, the stage one is MAE pretraining for conditioning utilising masks from other methods and MRF block. The obtained results are utilised in stage two with PBE base is used to train the ControlNet with conditioning inputs resulting in $x_{g}$ output image}
    \vspace{-0.05cm}
    \label{fig:architecture}
    \vspace{-0.25cm}
\end{figure*}

\section{Proposed Methodology}


We introduce DiffSTR, a diffusion based method that addresses the problem of scene text removal as an inpainting task. We use a pretrained PBE(Paint-by-diffusion) model\cite{yang2023paint} as our base diffusion model, which utilises a Latent Diffusion Model (LDM)\cite{Rombach_2022_CVPR} for image to image generation. The model is pre-trained on the task of image inpainting based on the exemplar image provided\cite{yang2023paint}. Training such a heavy model for the task of scene text removal is computationally intensive and requires loads of training data.  We introduce ControlNet to the PBE diffusion model, that is conditioned on coarse textless image generated from a masked autoencoder (TR-MAE) as shown in Figure \ref{fig:architecture}. The MAE is trained on textless images. 
We also propose a segmentation mask refinement framework that iteratively refines an initial mask and segments it using the SLIC\cite{6205760} and Hierarchical Feature Selection (HFS)\cite{cheng2016hfs} algorithms to produce an accurate final text mask.   

\textbf{Architecture flow}
Our approach comprises of two stages. In stage 1, our novel segmentation-based mask refinement framework(MRF) predicts accurate text mask $y_{MRF}$ while taking SAEN stroke mask $y_{SAEN}$ as input. The need arises to accurately encompass the text surrounding area.
The resulting mask $y_{MRF}$ is utilised in TR-MAE block with input masks from other approaches like PERT, CTRnet, etc. to train a masked autoencoder (MAE) on scene images not containing any text. This helps MAE to learn to predict a coarse textless image $x_{c}$ for our stage 2. 

The stage 2 employs a ControlNet architecture\cite{zhang2023adding}, which introduces additional control conditions while preserving the generative capabilities of PBE. The PBE model is kept frozen as shown in Figure \ref{fig:architecture}.  The PBE model gets the following inputs while training: input image $x_{0}$, mask image $y_{MRF}$ and reference input image $x_{i}$. The PBE model utilises CLIP image encoder's embeddings extracted from the coarse input $x_{c}$ received from TR-MAE for conditioning. The mask image $y_{MRF}$ is multiplied by the input scene image $x_{c}$ to compute an image $x_h$ that only contains the coarse region to be inpainted. 

The ControlNet encoder gets all the inputs of the PBE model, along with features computed from $x_h$ as an additional input as shown in Figure \ref{fig:architecture}. DINO-V2 image encoder's embeddings $f_{c}$ are extracted from the coarse image $x_{c}$. The embeddings $f_{c}$ are then sent to a learnable linear layer and then fed to the ControlNet. The ControlNet learns to condition the PBE model to generate textless image $x_g$. Here, we describe our proposed approach DiffSTR in detail.

\subsection{Diffusion for Text removal}


\subsubsection{ControlNet}

We approach the scene test removal problem as solving an image inpainting task utilising PBE model\cite{yang2023paint}, which is an image conditioned diffusion model.  
\textbf{The PBE model} alters the masked image region semantically, given the exemplar image, which makes it robust for inpainting tasks. The PBE model uses Stable Diffusion\cite{Rombach_2022_CVPR} as the Latent Diffusion Model. The model utilises Denoising Unet architecture comprising of multiple standard diffusion Encoder blocks, multiple Decoder blocks and one middle block. The Decoder and Encoder blocks are connected via skip connections. 
Though the results generated by pretrained PBE model are semantically consistent with the input reference image, the generated image is not  structurally consistent (For  example, a dog can be inpainted, but it's facial features, shape and surroundings may change.) Text removal requires that the inpainted region is both structurally and semantically consistent to the coarse reference image. Thus, there's a need to add structural control to the existing PBE model. To introduce the same, our proposed approach conditions the pretrained PBE (Paint By Example) model by introducing a trainable ControlNet to provide controlled learning of the scene text removal task without the  computational overhead of training entire PBE. 


The DiffSTR stage 2 takes as inputs the input image \(x_{0}\), binary mask \(y_{MRF}\), reference input image \(x_{i}\), timesteps \(t\), and additional conditional inputs \(x_{c}\) and \(x_{h}\). These inputs are used for the generation of control vectors, which go through zero convolution layers before being fed to the PBE decoder and middle block to guide the generation.

To provide a better texture regeneration, the ControlNet encoder is also provided with conditioning inputs $f_{c}$ from Dino-v2, $x_{c}$ and $x_{h}$ (obtained from MRF and TR-MAE) explained in the further sections. 

The input image $x_{0}$ and reference image $x_{i}$ are first sent as an input to VAE encoder(s)\cite{Rombach_2022_CVPR} which compute latents $z$ and $z_{i}$ respectively. Noise at timestep 't' is added to $z$ to compute $z_t$. The segmentation mask $y_{MRF}$ is downsampled 4 times and concatenated with the latents $z_{t}$ and $z_{i}$. The same are fed to a convolutional layer whose output serves as input to the denoising Unet. The denoising unet learns the noise added to the latent $z$. The output latent is fed to the VAE decoder to generate the textless image $x_{g}$.


The stage 2 block utilises the above inputs and is trained with the encoder and middle block parameters of PBE copied to the ControlNet encoder with the training process updating only the gradients of the parameters of the ControlNet encoder exclusively.

The DiffSTR module learns the noise added to the image latent $z$ by learning a network $\epsilon_{\theta}$ minimising the following objective function:
\vspace{-0.3cm}
\begin{equation}
    \vspace{-0.6cm}
    \mathbb{E}_{\encoder{(\mathbf{x}_0)},\encoder{(\mathbf{x}_i)}, \mathbf{y}_{MRF},\mathbf{x}_h ,t,\epsilon}\left[\left \Vert \epsilon-\epsilon_\theta(\mathbf{z},\mathbf{z}_i, \mathbf{y}_{MRF},\mathbf{x}_h ,t) \right \Vert^2_2\right]
    \label{noiseeq}
\end{equation}


\subsubsection{ControlNet Conditioning}
The PBE method shows capability in the generation of realistic images but suffers from pixel-level control which often results in an inaccurate reconstruction of patterns and textures of background scene in images. The reason for this problem is the usage of CLIP architecture as an encoder to extract the features from the conditional inputs. Although CLIP\cite{radford2021learning} performs well to align the image in other tasks, it faces issues in scene text removal. The scene text background contains varied textures which makes it challenging to restore the same while removing the text. The semantic information from CLIP lacks sufficient features to accurately define the patterns and textures of the background.

In order to provide a texture level consistency to the generated scene, a pretrained DINO-V2\cite{oquab2024dinov2learningrobustvisual} Image Encoder is employed. In DINO-V2, the images are encoded as global tokens as well as patch tokens. Such local details lead to the preservation of more pixel-level information along with global information in the conditioning image. 
DINO-V2 takes in coarse image $x_{c}$ as the conditional input from TR-MAE and computes embedding $f_{c}$ as shown in Figure \ref{fig:architecture}. The features $f_{c}$ from DINO-V2 are passed through a linear layer and integrated into UNet through a standard cross attention  mechanism\cite{10.5555/3295222.3295349}, thereby increasing the pixel controllability in DiffSTR significantly.


\subsection{Segmentation based mask refinement framework (MRF)}
In the scene text removal task using image-to-image latent diffusion models, inpainting is employed to generate text-free images. The inpainting mask identifies text regions, guiding the model to accurately remove text. Nevertheless, the efficacy of text erasure is significantly influenced by the mask's precision. Inaccurate or insufficient mask predictions can cause distorted scene reconstructions, leading to poor text removal.
One stage solutions like PERT have integrated mask prediction into the text removal step. Though this eliminates the computation of masked image, the trained model still depends heavily on the mask training data. We wish to disentangle the mask prediction from the training data, by incorporating a novel mask refinement framework that gives an accurate inpainting mask with less sensitivity to the training data. 

Though training a deep learning model is an optimal way to do text stroke segmentation, the dependence on correctness of ground truth masks is a necessary yet difficult task. One solution is to train the text stroke segmentation on multiple datasets, similar to ViTeraser\cite{peng2024viteraser}. Though it improves the robustness, it doesn't eliminate the dependence on the ground truth stroke mask. Our model utilises masks predicted by SAEN as an initial mask. The initial mask is then upsampled, eroded and downsampled iteratively, to compute an initial seed mask for segmentation. The seed mask is fed to the SLIC\cite{6205760} algorithm to compute small superpixel regions which is then fed to the Hierarchical Feature Selection (HFS)\cite{cheng2016hfs} for computing final segmented text mask $y_{MRF}$. 
\begin{figure}
    \centering
    \includegraphics[width=1\linewidth]{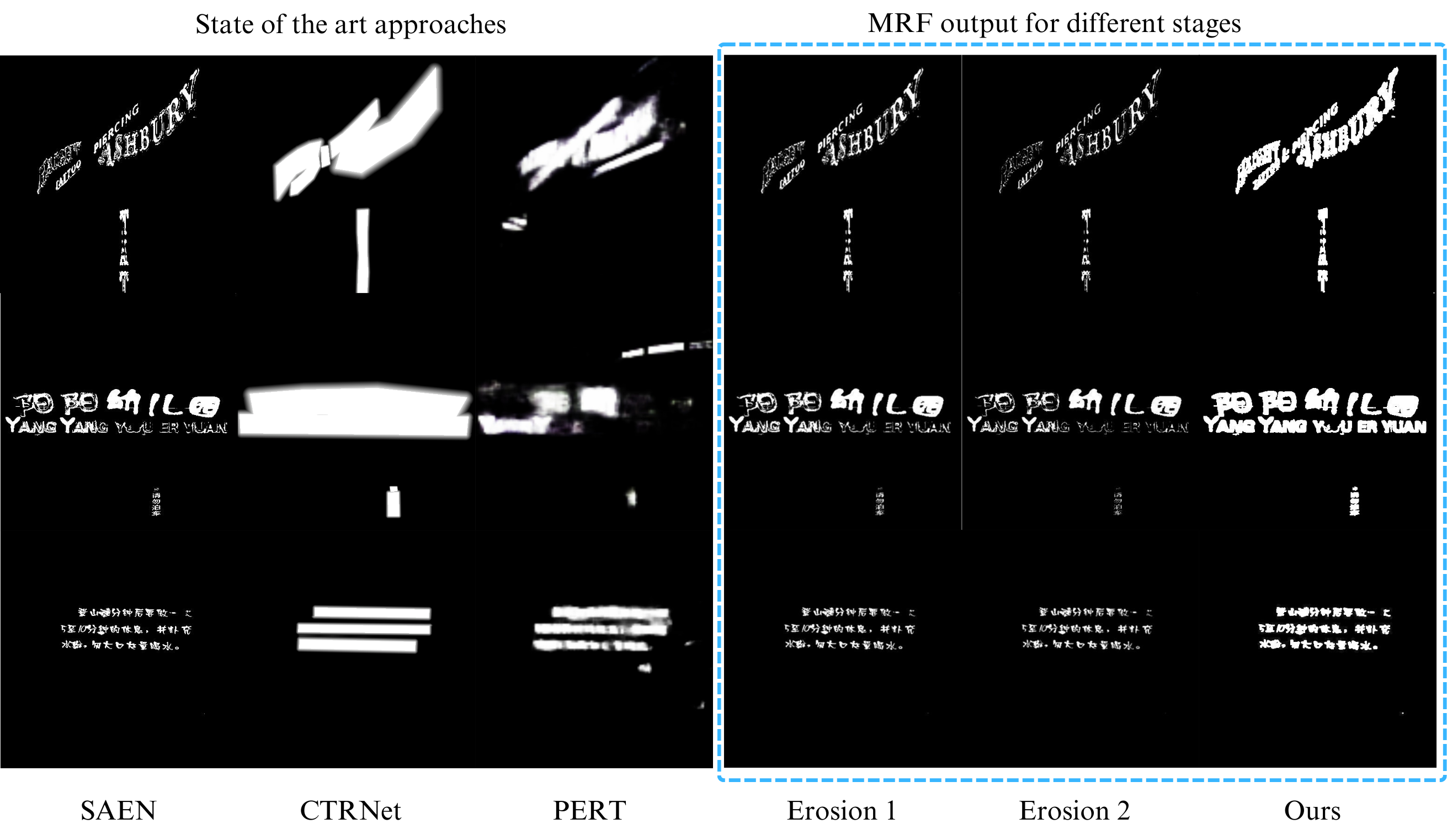}
    \caption{Qualitative results and ablation of segmentation masks generated by our novel MRF module. The left three columns show the SOTA masking approaches while the blue box on the right shows ablation and final results of MRF module}
    \label{fig:qualitative_mrf}
    \vspace{-0.25cm}
\end{figure}

\vspace{-0.2cm}
\subsubsection{Masked pretraining for conditioning (TR-MAE)}

Diffusion models rely on predicting random noise for diverse generation, which often results in generation of random non-coherent objects within image. Utilizing the diffusion model for text removal is one such problem. This highlights the need for conditional image generation. In context of text removal, the condition/control should comprise of image/features describing region around the foreground text to be removed. MAE trained on textless images provides one such context for a prediction that is reliant solely on the unmasked regions of the image, thereby giving a coarse textless image $x_c$. 
The binary mask input is multiplied with input image $x_0$ resulting in the reference masked image $x_i$ which is fed to the MAE for training. The MAE is trained on an L2 loss between $x_{gt}$ and $x_c$ where $x_{gt}$ is the textless ground truth image. However, training MAE using only random masks and augmentation is sub-optimal for learning to remove textual regions from the image. To deal with this, we employ the training of MAE using masks present in the text region of the corresponding image containing text.
These binary masks are a combination of a set of masks computed from PERT, SAEN, CTRNet and the mask generated from our novel segmentation-based mask refinement framework(MRF). This ensures training of MAE on box masks, coarse stroke masks and detailed stroke masks. We keep the ratio of masks as 20 percent, 30 percent and 50 percent, respectively, while training. Training MAE on such masks makes it more robust for the task of text removal.








\section{Experiments}
\subsection{Datasets}
\begin{figure*}
    \centering
    \includegraphics[width=0.8\linewidth]{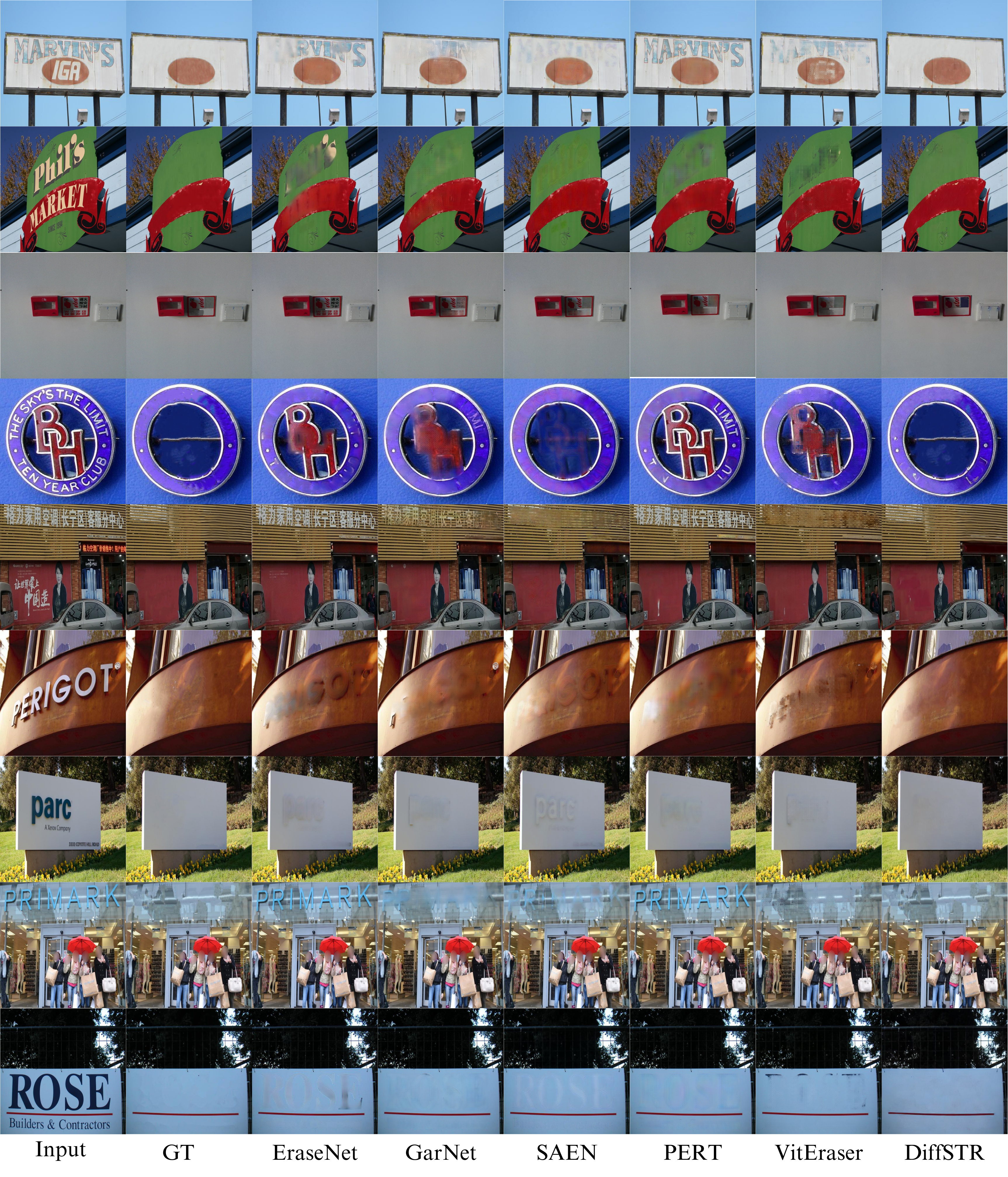}
    \vspace{-0.20cm}
    \caption{Qualitative comparison of our proposed scene text removal framework with state-of-the-art on SCUT-EnsText dataset.}
    \vspace{-0.35cm}
    \label{fig:qualitative_ens}
    \vspace{-0.15cm}
\end{figure*}

WE trained our model on two datasets, SCUT-EnsText\cite{Erase} and SCUT-Syn\cite{Erase}. SCUT-EnsText is a real world dataset with 2,749 samples for training and 813 samples for testing, which are selected from public scene text detection benchmarks. The dataset presents diverse text types, including text fonts, text shapes, text orientations and scene diversity. The ground-truth is computed by manually erasing all text instances using Adobe Photoshop. SCUT-Syn dataset is a synthetic dataset 8,000 and 800 samples for training and testing, respectively. The SCUT-Syn dataset is prepared with text synthesis where the background images are collected from  ICDAR 2013 and ICDAR MLT-2017 and the texts are manually erased. The dataset consists of scene images with text and corresponding manually computed ground truth serving as the prediction target, examples can be seen from Figure \ref{fig:qualitative_ens}. 

\subsection{Implementation Details}
During the experiments, we used an end-to-end training process. All experiments were conducted using a single A100 GPU (80 GB) with image resolutions of \(512 \times 512\). We used the AdamW optimizer with a learning rate of \(2 \times 10^{-5}\) and  1000 timesteps for training the diffusion model.

\textbf{Training MAE} For the MAE \cite{he2022masked} used in our proposed TR-MAE module, we train on textless images of size 256x256, which is efficient and produces context-stable guidance for SD to generate high-resolution textless images. We fine-tune from the pre-trained MAE with a batch size of 1024. We use AdamW \cite{loshchilov2017decoupled} optimizer with learning rate of 5e-2. We also incorporate ColorJitter for color augmentation, with brightness 0.15, contrast 0.2, saturation 0.1, and hue 0.03.

\begin{table*}[t]
    \centering 
    \normalsize
    \setlength{\tabcolsep}{1.5mm}
    \resizebox{1.75\columnwidth}{!}{
        \begin{tabular}{llccccccccccccc}
        \toprule 
        \multirow{2}*{Method} & \multirow{2}*{Venue} & \multicolumn{6}{c}{SCUT-EnsText} & \multicolumn{6}{c}{SCUT-Syn} & \multirow{2}*{Speed(fps)}\\
        \cmidrule(r){3-8} \cmidrule(r){9-14}
         &  & PSNR$\uparrow$ & MSSIM$\uparrow$ & MSE$\downarrow$ & AGE$\downarrow$ & pEPs$\downarrow$ & pCEPs$\downarrow$ & PSNR$\uparrow$ & MSSIM$\uparrow$ & MSE$\downarrow$ & AGE$\downarrow$ & pEPs$\downarrow$ & pCEPs$\downarrow$  \\
        \midrule
        Pix2pix \cite{isola2017image} & CVPR'17 & 26.70 & 88.56 & 0.37 & 6.09 & 0.0480 & 0.0227 & 26.76 & 91.08 & 0.27 & 5.47 & 0.0473 & 0.0244 & 133 \\
        STE \cite{nakamura2017scene} &  ICDAR'17 & 25.47 & 90.14 & 0.47 & 6.01 & 0.0533 & 0.0296 & 25.40 & 90.12 & 0.65 & 9.49 & 0.0553 & 0.0347 & - \\
        EnsNet \cite{zhang2019ensnet} & AAAI'19 & 29.54 & 92.74 & 0.24 & 4.16 & 0.0307 & 0.0136 & 37.36 & 96.44 & 0.21 & 1.73 & 0.0069 & 0.0020 & 199\\
        MTRNet++ \cite{tursun2020mtrnet++} & CVIU'20 & 29.63 & 93.71 & 0.28 & 3.51 & 0.0305 & 0.0168 & 34.55 & 98.45 & 0.04 & - & - & - & 53 \\
        EraseNet \cite{liu2020erasenet} & TIP'20 & 32.30 & 95.42 & 0.15 & 3.02 & 0.0160 & 0.0090 & 38.32 & 97.67 & 0.02 & 1.60 & 0.0048 & 0.0004 & 71 \\
        STRDD \cite{strdd} & ISAIR'22 & 32.60 & 93.80 & 0.15 & - & - & - & - & - & - & - & - & - & 0.62 \\
        SSTE \cite{Tang_2021} & TIP'21 & 35.34 & 96.24 & 0.09 & - & - & - & 38.60 & 97.55 & 0.02 & - & - & - & 7.8\\ 
        PSSTRNet \cite{Lyu_2022} & ICME'22 & 34.65 & 96.75 & 0.14 & 1.72 & 0.0135 & 0.0074 & 39.25 & 98.15 & 0.02 & 1.20 & 0.0043 & 0.0008 & 56\\
        CTRNet \cite{CTRNet} & ECCV'22 & 35.20 & 97.36 & 0.09 & 2.20 & 0.0106 & 0.0068 & 41.28 & 98.52 & 0.02 & 1.33 & 0.0030 & 0.0007 & 5.1\\
        GaRNet$\S$ \cite{lee2022surprisingly} & ECCV'22 & 35.45 & 97.14 & 0.08 & 1.90 & 0.0105 & 0.0062 &  & - & - & - & - & - & 22\\
        MBE \cite{Hou_2022_ACCV} & ACCV'22 & 35.03 & 97.31 & - & 2.06 & 0.0128 & 0.0088 & 43.85 & 98.64 & - & 0.94 & 0.0013 & 0.0004 & - \\
        PEN \cite{du2023progressivescenetexterasing} & CVIU'23 & 35.21 & 96.32 & 0.08 & 2.14 & 0.0097 & 0.0037 & 39.26 & 98.03 & 0.02 & 1.29 & 0.0038 & 0.0004 & - \\
        PEN* \cite{du2023progressivescenetexterasing} & CVIU'23 & 35.72 & 96.68 & 0.05 & 1.95 & 0.0071 & 0.0020 & 38.87 & 97.83 & 0.03 & 1.38 & 0.0041 & 0.0004 & - \\
        PERT \cite{wang2021pert} & TIP'23 & 33.62 & 97.00 & 0.13 & 2.19 & 0.0135 & 0.0088 & 39.40 & 97.87 & 0.02 & 1.41 & 0.0046 & 0.0007 & 24 \\ 
        SAEN \cite{du2023modeling} & WACV'23 & 34.75 & 96.53 & 0.07 & 1.98 & 0.0125 & 0.0073 & 38.63 & 98.27 & 0.03 & 1.39 & 0.0043 & 0.0004 & 62\\
        FETNet \cite{LYU2023109531} & PR'23 & 34.53 & 97.01 & 0.13 & 1.75 & 0.0137 & 0.0080 & 39.14 & 97.97 & 0.02 & 1.26 & 0.0046 & 0.0008 & 77 \\  
        ViTEraser-Base\cite{peng2024viteraser} & AAAI'24 & 36.32 & 97.51 & 0.05 & 1.86 & 0.0074 & 0.0041 & 42.53 & 98.45 & 0.01 & 1.19 & 0.0018 & 0.00016 & 15 \\ 
        \midrule
        DiffSTR & - & \textbf{37.25}          & \textbf{97.98}           & \textbf{0.04}          & \textbf{1.61}        & \textbf{0.0071}        & \textbf{0.0038} & \textbf{43.75}          & \textbf{98.83}           & \textbf{0.01}          & \textbf{1.12}        & \textbf{0.0013}        & \textbf{0.00014} & 2
        \underline{} \\ 
        \bottomrule
    \end{tabular}}
    \vspace{-0.2cm}
    \caption{Comparison with state of the arts on SCUT-EnsText and SCUT-Syn. (Bold: state of the art, underline: the second best)}
    \vspace{-0.1cm}
    \label{tab:exp_scut_enstext}
\end{table*}

\subsection{Quantitative Performance evaluation}
We compare our proposed method with the existing state-of-the-art approaches for scene text removal. The evaluation for all the methods mentioned in Table \ref{tab:exp_scut_enstext} has been done on SCUT-EnsText and SCUT-Syn datasets. The inference was done on a batch size 1 for all the methods mentioned.
Although our text mask was not trained on any auxilary dataset like in \cite{peng2024viteraser}, yet the quantitative results demonstrate that our novel segmentation based mask refinement framework(MRF), along with introduction of ControlNet diffusion model conditioned on a coarse textless image, generated by TR-MAE, significantly improves the generated image, thereby getting our proposed approach the state-of-the-art performance on real world scene text removal. This confirms the effectiveness of DiffSTR in real world scenarios.
Since our proposed method considers the text removal as a generative inpainting task, we have employed the inpainting metrics for evaluation of DiffSTR, similar to other popular approaches\cite{du2023modeling}. 
The proposed approach shows significant improvements over previous methods by improving the PSNR from  36.32dB to 37.25dB which reflects on the effectiveness of our method in terms of efficient generation. The high values achieved in MSSIM shows the robust structural consistency achieved in the generated images. The metric AGE, pEPs and pCEPs evaluates the generated images at gray absolute pixel difference, the percentage of pixel errors and the error in terms of pixel neighbour respectively. This significant reduction of the metrics as shown in Table \ref{tab:exp_scut_enstext} shows the effectiveness of our approach in scene text removal while keeping the structure, semantics and texture attributes intact.
While our model runs at 2 fps, it is significantly better than STRDD( The other diffusion based approach), in terms of both accuracy and run time complexity. Furthermore, approaches like InstaFlow\cite{liu2023instaflow} and Block caching\cite{wimbauer2024cache} can improve the run time of proposed method significantly for real world usage.

\begin{figure}
    \centering
    \includegraphics[width=1\linewidth]{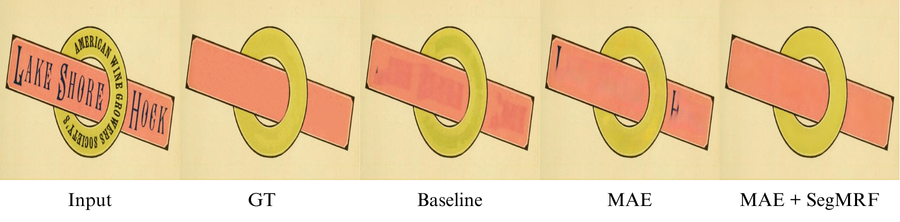}
    \vspace{-0.6cm}
    \caption{Qualitative comparison of ablation on three modules (diffusion baseline , TR-MAE and MRF)}
    \label{fig:qualitative_ablation}
    \vspace{-0.35cm}

\end{figure}

\subsection{Qualitative comparison with benchmarks}
As shown in Figure \ref{fig:qualitative_ens}, row 1 demonstrates our method's ability to completely erase text from an image with a patched texture, which can be ambiguous for methods that rely heavily on stroke patterns. DiffSTR preserves the complex background texture while removing the text, surpassing all existing methods, as shown in row 4. The results in row 5 highlight the perfect preservation of complex background textures by our method. Text removal becomes challenging on white backgrounds due to high-contrast replacement, creating text ghosting artifacts, as shown in rows 7 and 9. Our method effectively solves this issue. These improvements are due to our proposed diffusion approach that's conditioned on outputs from our trained masked autoencoder, fine-tuned with accurate masks to generate a coarse, precise conditional image.

In real-world scenarios, text masks are often poorly defined, leading to blurred textures in the background, as seen with other methods\cite{wang2021pert}. This issue is resolved by our segmentation-based mask refinement framework, as depicted in rows 2 and 6. Row 3 shows that only our method effectively handles small text conditions without leaving mask patches or text remnants, as seen in the red box text removal. Text removal on transparent and reflective surfaces, as shown in row 8, can be challenging to retain the layered background, where DiffSTR shows remarkable success.

\subsubsection{Qualitative analysis of MRF block}
The Figure \ref{fig:qualitative_mrf} shows the qualitative comparison between the text mask generated from our novel MRF block in comparison to the state-of-the-art methods. As can be seen from the figure, our final mask is able to segment-out more text in the image than any other approach. Our method also generates slightly dialated masks which prevents boundary artefacts due to inpainting. The blue box shows a visual ablation of generated mask for every stage of MRF module. The final resulting text mask utilises erosion2 as seed mask and shows significantly improved results.


\begin{figure}
    \centering
    \includegraphics[width=\linewidth]{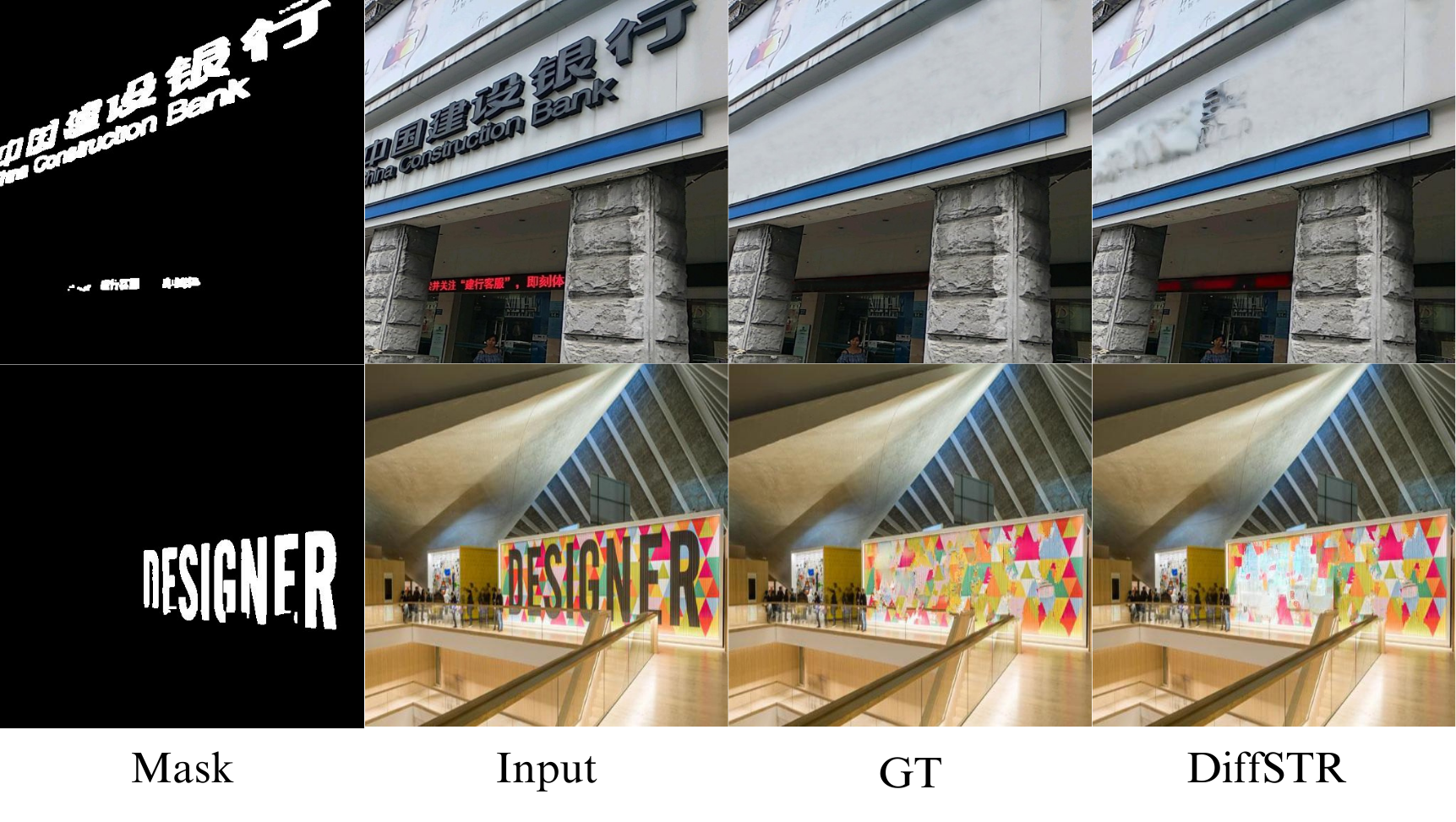}
    \vspace{-0.25cm}
    \vspace{-0.5cm}
    \caption{Failure cases of DiffSTR}
    \label{fig:qual_limitation}
    \vspace{-0.4cm}
\end{figure}

\begin{table*}[ht]
\centering
    \normalsize
\begin{tabular}{lllrrrrrr}
\toprule
ControlNet & TR-MAE & MRF & \multicolumn{1}{l}{PSNR} & \multicolumn{1}{l}{MSSIM} & \multicolumn{1}{l}{MSE} & \multicolumn{1}{l}{AGE} & \multicolumn{1}{l}{pEPs} & \multicolumn{1}{l}{pCEPs} \\
\midrule
\xmark          & \xmark   & \xmark      & 33.169                   & 0.9536                    & 0.15                    & 2.681                   & 0.0152                   & 0.00919                 \\
\checkmark          & \xmark   & \xmark      & 34.122                   & 0.9563                    & 0.14                    & 2.571                   & 0.0144                   & 0.00832                   \\
\checkmark          & \xmark   & \checkmark      & 35.290                    & 0.9676                    & 0.12                    & 2.178                   & 0.0126                  & 0.00783                   \\
\checkmark          & \checkmark   & \xmark      & 36.268                   & 0.9687                    & 0.08                    & 1.959                  & 0.0119                   & 0.00727  \\
\bottomrule
\checkmark          & \checkmark   & \checkmark      & \textbf{37.257}          & \textbf{0.9798}           & \textbf{0.04}          & \textbf{1.612}        & \textbf{0.0071}        & \textbf{0.00383}

\end{tabular}
\caption{Ablation study of the proposed methodology}
    \label{tab:ablation}
\end{table*}

\begin{table*}[ht]
\centering
    \normalsize
\begin{tabular}{llrrrrrr}
\toprule
Initial Mask      & MRF & \multicolumn{1}{l}{PSNR} & \multicolumn{1}{l}{MSSIM} & \multicolumn{1}{l}{MSE} & \multicolumn{1}{l}{AGE} & \multicolumn{1}{l}{pEPs} & \multicolumn{1}{l}{pCEPs} \\
\midrule
CTRNet(Box mask)  & \xmark      & 35.805                   & 0.9670                     & 0.09                    & 2.0412                  & 0.0128                   & 0.00792                   \\
CTRNet(Box mask)  & \checkmark      & 36.641                   & 0.9694                    & 0.07                    & 1.8652                  & 0.0107                  & 0.00646                   \\
SAEN(Stroke mask) & \xmark      & 36.684                   & 0.9717                  & 0.07                    & 1.6997                  & 0.0083                 & 0.00493                  \\
\bottomrule
SAEN(Stroke mask) & \checkmark      & \textbf{37.257}          & \textbf{0.9798}           & \textbf{0.04}          & \textbf{1.6122}        & \textbf{0.0071}        & \textbf{0.00383}         
\end{tabular}
\caption{Ablation study of our Segmentation based mask refinement framework}
    \label{tab:ablation_detailed}
    \vspace{-0.25cm}
\end{table*}

\subsection{Ablation study and Analysis}
To understand the impact of each component in our framework, we conducted an ablation study by systematically removing or modifying one module at a time and observing the effects on performance.
We first evaluated the contribution of the ControlNet diffusion model by comparing it with a pretrained PBE model as a baseline. The results, presented in Table \ref{tab:ablation}, show that the diffusion model significantly improves both PSNR and SSIM scores, indicating better fidelity and texture consistency. 

Next, we examined the effectiveness of the mask pretraining pipeline. We compared the ControlNet with an inpainting input image as a condition to the coarse textless image generated by our fine-tuned masked autoencoder (MAE). As shown in Table \ref{tab:ablation} and Figure \ref{fig:qualitative_ablation}, the fine-tuned MAE considerably enhances the model's robustness to text removal, reflected in improved PSNR and SSIM scores.

Finally, we assessed the contribution of our segmentation-based mask refinement framework by integrating it into our pipeline and comparing the results with those obtained using initial masks from CTRNet and SAEN alone. The integration of the refinement framework with text stroke masks from SAEN as the initial seed resulted in significant gains in PSNR, SSIM, and FID. Overall, the results of our experiments and ablation study highlight the effectiveness of each component in our proposed method and underscore its ability to outperform existing state-of-the-art approaches in scene text removal.

\textbf{Ablation of MRF module:} 
To isolate the contribution of the MRF framework, we evaluated the performance of our models with and without MRF module. We used two common scene text mask predictors, CTRNet and SAEN, along with their respective mask types (box mask for CTRNet and stroke mask for SAEN). As shown in Table \ref{tab:ablation_detailed}, inclusion of the MRF framework significantly enhances performance across all metrics. For instance, when using CTRNet with a box mask, MRF improves PSNR from 35.805 to 36.641 and MSSIM from 0.9670 to 0.9694, while also reducing MSE, AGE, pEPs, and pCEPs. Similarly, with SAEN using a stroke mask, MRF boosts PSNR from 36.684 to 37.257 and MSSIM from 0.9717 to 0.9798, with corresponding improvements in other metrics. These results highlight that MRF effectively addresses mask inaccuracy, leading to more precise text removal and consequently better scene image quality.




\subsection{Limitations}
The dependence of the diffusion model on high-quality initial masks makes it dependent on the quality of the mask and thus given an incomplete text mask, it is unable to remove text completely. Slight variation in color-contrast consistency of generated image can be seen in row(2) of Figure \ref{fig:qual_limitation} due to presence of multiple complex colors and textures. This can be improved by more availability of training data with complex textures and multiple colors.


\section{Conclusion}
In this work, we introduced a novel approach to Scene Text Removal (STR) using the advanced generative capabilities of latent diffusion models. By leveraging a ControlNet diffusion model and treating STR as a conditional inpainting task with ControlNet diffusion model, we tackled key challenges in achieving high-fidelity, texture-consistent scene generation. Our method incorporated mask pretraining pipeline, where a masked autoencoder (MAE) was fine-tuned using the masks derived from our innovative segmentation-based mask refinement framework. This framework enhanced the accuracy of inpainting masks by generating precise and contextually aware masks, reducing reliance on training data. Extensive experiments on the SCUT-EnsText and SCUT-Syn datasets demonstrated that our approach significantly outperforms existing state-of-the-art methods, producing visually compelling textless images with minimal artifacts. Our work highlights the potential of diffusion models in complex image editing tasks and sets a new benchmark in STR, with promising future directions in further optimizing the model architecture and extending the approach to other sensitive information removal and general image inpainting tasks.



{\small
\bibliographystyle{ieee_fullname}
\bibliography{egbib}

\begin{thebibliography}{10}\itemsep=-1pt

\bibitem{6205760}
Radhakrishna Achanta, Appu Shaji, Kevin Smith, Aurelien Lucchi, Pascal Fua, and Sabine Süsstrunk.
\newblock Slic superpixels compared to state-of-the-art superpixel methods.
\newblock {\em IEEE Transactions on Pattern Analysis and Machine Intelligence}, 34(11):2274--2282, 2012.

\bibitem{avrahami2022blended}
Omri Avrahami, Dani Lischinski, and Ohad Fried.
\newblock Blended diffusion for text-driven editing of natural images.
\newblock In {\em Proceedings of the IEEE/CVF conference on computer vision and pattern recognition}, pages 18208--18218, 2022.

\bibitem{barnes2009patchmatch}
Connelly Barnes, Eli Shechtman, Adam Finkelstein, and Dan~B Goldman.
\newblock Patchmatch: A randomized correspondence algorithm for structural image editing.
\newblock {\em ACM Trans. Graph.}, 28(3):24, 2009.

\bibitem{bertalmio2000image}
Marcelo Bertalmio, Guillermo Sapiro, Vincent Caselles, and Coloma Ballester.
\newblock Image inpainting.
\newblock In {\em Proceedings of the 27th annual conference on Computer graphics and interactive techniques}, pages 417--424, 2000.

\bibitem{bian2022scene}
Xuewei Bian, Chaoqun Wang, Weize Quan, Juntao Ye, Xiaopeng Zhang, and Dong-Ming Yan.
\newblock Scene text removal via cascaded text stroke detection and erasing.
\newblock {\em Computational Visual Media}, 8:273--287, 2022.

\bibitem{cao2022learning}
Chenjie Cao, Qiaole Dong, and Yanwei Fu.
\newblock Learning prior feature and attention enhanced image inpainting.
\newblock In {\em European conference on computer vision}, pages 306--322. Springer, 2022.

\bibitem{chen2024context}
Xiaokang Chen, Mingyu Ding, Xiaodi Wang, Ying Xin, Shentong Mo, Yunhao Wang, Shumin Han, Ping Luo, Gang Zeng, and Jingdong Wang.
\newblock Context autoencoder for self-supervised representation learning.
\newblock {\em International Journal of Computer Vision}, 132(1):208--223, 2024.

\bibitem{cheng2016hfs}
Ming-Ming Cheng, Yun Liu, Qibin Hou, Jiawang Bian, Philip Torr, Shi-Min Hu, and Zhuowen Tu.
\newblock Hfs: Hierarchical feature selection for efficient image segmentation.
\newblock In {\em Computer Vision--ECCV 2016: 14th European Conference, Amsterdam, The Netherlands, October 11-14, 2016, Proceedings, Part III 14}, pages 867--882. Springer, 2016.

\bibitem{chu2023rethinking}
Tianyi Chu, Jiafu Chen, Jiakai Sun, Shuobin Lian, Zhizhong Wang, Zhiwen Zuo, Lei Zhao, Wei Xing, and Dongming Lu.
\newblock Rethinking fast fourier convolution in image inpainting.
\newblock In {\em Proceedings of the IEEE/CVF International Conference on Computer Vision}, pages 23195--23205, 2023.

\bibitem{conrad2021two}
Benjamin Conrad and Pei-I Chen.
\newblock Two-stage seamless text erasing on real-world scene images.
\newblock In {\em 2021 IEEE International Conference on Image Processing (ICIP)}, pages 1309--1313. IEEE, 2021.

\bibitem{du2023modeling}
Xiangcheng Du, Zhao Zhou, Yingbin Zheng, Tianlong Ma, Xingjiao Wu, and Cheng Jin.
\newblock Modeling stroke mask for end-to-end text erasing.
\newblock In {\em Proceedings of the IEEE/CVF Winter Conference on Applications of Computer Vision}, pages 6151--6159, 2023.

\bibitem{du2023progressivescenetexterasing}
Xiangcheng Du, Zhao Zhou, Yingbin Zheng, Xingjiao Wu, Tianlong Ma, and Cheng Jin.
\newblock Progressive scene text erasing with self-supervision, 2023.

\bibitem{epshtein2010detecting}
Boris Epshtein, Eyal Ofek, and Yonatan Wexler.
\newblock Detecting text in natural scenes with stroke width transform.
\newblock In {\em 2010 IEEE computer society conference on computer vision and pattern recognition}, pages 2963--2970. IEEE, 2010.

\bibitem{goodfellow2014generative}
Ian Goodfellow, Jean Pouget-Abadie, Mehdi Mirza, Bing Xu, David Warde-Farley, Sherjil Ozair, Aaron Courville, and Yoshua Bengio.
\newblock Generative adversarial nets.
\newblock {\em Advances in neural information processing systems}, 27, 2014.

\bibitem{he2022masked}
Kaiming He, Xinlei Chen, Saining Xie, Yanghao Li, Piotr Doll{\'a}r, and Ross Girshick.
\newblock Masked autoencoders are scalable vision learners.
\newblock In {\em Proceedings of the IEEE/CVF conference on computer vision and pattern recognition}, pages 16000--16009, 2022.

\bibitem{ho2020denoising}
Jonathan Ho, Ajay Jain, and Pieter Abbeel.
\newblock Denoising diffusion probabilistic models.
\newblock {\em Advances in neural information processing systems}, 33:6840--6851, 2020.

\bibitem{Hou_2022_ACCV}
Yujie Hou, Jiwei~Ji Chen, and Zengfu Wang.
\newblock Multi-branch network with ensemble learning for text removal in the wild.
\newblock In {\em Proceedings of the Asian Conference on Computer Vision (ACCV)}, pages 1333--1349, December 2022.

\bibitem{huang2014robust}
Weilin Huang, Yu Qiao, and Xiaoou Tang.
\newblock Robust scene text detection with convolution neural network induced mser trees.
\newblock In {\em Computer Vision--ECCV 2014: 13th European Conference, Zurich, Switzerland, September 6-12, 2014, Proceedings, Part IV 13}, pages 497--511. Springer, 2014.

\bibitem{inai2014selective}
Kohei Inai, M{\aa}rten P{\aa}lsson, Volkmar Frinken, Yaokai Feng, and Seiichi Uchida.
\newblock Selective concealment of characters for privacy protection.
\newblock In {\em 2014 22nd International Conference on Pattern Recognition}, pages 333--338. IEEE, 2014.

\bibitem{isola2017image}
Phillip Isola, Jun-Yan Zhu, Tinghui Zhou, and Alexei~A Efros.
\newblock Image-to-image translation with conditional adversarial networks.
\newblock In {\em Computer Vision and Pattern Recognition (CVPR), 2017 IEEE Conference on}, 2017.

\bibitem{kawar2023imagic}
Bahjat Kawar, Shiran Zada, Oran Lang, Omer Tov, Huiwen Chang, Tali Dekel, Inbar Mosseri, and Michal Irani.
\newblock Imagic: Text-based real image editing with diffusion models.
\newblock In {\em Proceedings of the IEEE/CVF Conference on Computer Vision and Pattern Recognition}, pages 6007--6017, 2023.

\bibitem{keserwani2021text}
Prateek Keserwani and Partha~Pratim Roy.
\newblock Text region conditional generative adversarial network for text concealment in the wild.
\newblock {\em IEEE Transactions on Circuits and Systems for Video Technology}, 32(5):3152--3163, 2021.

\bibitem{kim2022diffusionclip}
Gwanghyun Kim, Taesung Kwon, and Jong~Chul Ye.
\newblock Diffusionclip: Text-guided diffusion models for robust image manipulation.
\newblock In {\em Proceedings of the IEEE/CVF conference on computer vision and pattern recognition}, pages 2426--2435, 2022.

\bibitem{ko2023continuously}
Keunsoo Ko and Chang-Su Kim.
\newblock Continuously masked transformer for image inpainting.
\newblock In {\em Proceedings of the IEEE/CVF International Conference on Computer Vision}, pages 13169--13178, 2023.

\bibitem{lee2022surprisingly}
Hyeonsu Lee and Chankyu Choi.
\newblock The surprisingly straightforward scene text removal method with gated attention and region of interest generation: A comprehensive prominent model analysis.
\newblock In {\em European Conference on Computer Vision}, pages 457--472. Springer, 2022.

\bibitem{CTRNet}
Chongyu Liu, Lianwen Jin, Yuliang Liu, canjie Luo, Bangdong Chen, Fengjun Guo, and Kai Ding.
\newblock Don’t forget me: Accurate background recovery for text removal via modeling local-global context.
\newblock {\em ECCV}, 2022.

\bibitem{liu2020erasenet}
Chongyu Liu, Yuliang Liu, Lianwen Jin, Shuaitao Zhang, Canjie Luo, and Yongpan Wang.
\newblock Erasenet: End-to-end text removal in the wild.
\newblock {\em IEEE Transactions on Image Processing}, 29:8760--8775, 2020.

\bibitem{Erase}
Chongyu Liu, Yuliang Liu, lianwen Jin, Shuaitao Zhang, Canjie Luo, and Yongpan Wang.
\newblock Erasenet: End-to-end text removal in the wild.
\newblock {\em IEEE Transactions on Image Processing}, 29:8760--8775, 2020.

\bibitem{liu2018image}
Guilin Liu, Fitsum~A Reda, Kevin~J Shih, Ting-Chun Wang, Andrew Tao, and Bryan Catanzaro.
\newblock Image inpainting for irregular holes using partial convolutions.
\newblock In {\em Proceedings of the European conference on computer vision (ECCV)}, pages 85--100, 2018.

\bibitem{liu2020textual}
Siqi Liu, Libiao Jin, and Fang Miao.
\newblock Textual restoration of occluded tibetan document pages based on side-enhanced u-net.
\newblock {\em Journal of Electronic Imaging}, 29(6):063006--063006, 2020.

\bibitem{liu2023instaflow}
Xingchao Liu, Xiwen Zhang, Jianzhu Ma, Jian Peng, and Qiang Liu.
\newblock Instaflow: One step is enough for high-quality diffusion-based text-to-image generation.
\newblock In {\em International Conference on Learning Representations}, 2024.

\bibitem{loshchilov2017decoupled}
Ilya Loshchilov and Frank Hutter.
\newblock Decoupled weight decay regularization.
\newblock {\em arXiv preprint arXiv:1711.05101}, 2017.

\bibitem{LYU2023109531}
Guangtao Lyu, Kun Liu, Anna Zhu, Seiichi Uchida, and Brian~Kenji Iwana.
\newblock Fetnet: Feature erasing and transferring network for scene text removal.
\newblock {\em Pattern Recognition}, 140:109531, 2023.

\bibitem{Lyu_2022}
Guangtao Lyu and Anna Zhu.
\newblock Psstrnet: Progressive segmentation-guided scene text removal network.
\newblock In {\em 2022 IEEE International Conference on Multimedia and Expo (ICME)}. IEEE, July 2022.

\bibitem{mokady2023null}
Ron Mokady, Amir Hertz, Kfir Aberman, Yael Pritch, and Daniel Cohen-Or.
\newblock Null-text inversion for editing real images using guided diffusion models.
\newblock In {\em Proceedings of the IEEE/CVF Conference on Computer Vision and Pattern Recognition}, pages 6038--6047, 2023.

\bibitem{nakamura2017scene}
Toshiki Nakamura, Anna Zhu, Keiji Yanai, and Seiichi Uchida.
\newblock Scene text eraser.
\newblock In {\em 2017 14th IAPR International Conference on Document Analysis and Recognition (ICDAR)}, volume~1, pages 832--837. IEEE, 2017.

\bibitem{nguyen2021dictionary}
Nguyen Nguyen, Thu Nguyen, Vinh Tran, Minh-Triet Tran, Thanh~Duc Ngo, Thien~Huu Nguyen, and Minh Hoai.
\newblock Dictionary-guided scene text recognition.
\newblock In {\em Proceedings of the IEEE/CVF Conference on Computer Vision and Pattern Recognition}, pages 7383--7392, 2021.

\bibitem{nichol2021glide}
Alex Nichol, Prafulla Dhariwal, Aditya Ramesh, Pranav Shyam, Pamela Mishkin, Bob McGrew, Ilya Sutskever, and Mark Chen.
\newblock Glide: Towards photorealistic image generation and editing with text-guided diffusion models.
\newblock {\em arXiv preprint arXiv:2112.10741}, 2021.

\bibitem{oquab2024dinov2learningrobustvisual}
Maxime Oquab, Timothée Darcet, Théo Moutakanni, Huy Vo, Marc Szafraniec, Vasil Khalidov, Pierre Fernandez, Daniel Haziza, Francisco Massa, Alaaeldin El-Nouby, Mahmoud Assran, Nicolas Ballas, Wojciech Galuba, Russell Howes, Po-Yao Huang, Shang-Wen Li, Ishan Misra, Michael Rabbat, Vasu Sharma, Gabriel Synnaeve, Hu Xu, Hervé Jegou, Julien Mairal, Patrick Labatut, Armand Joulin, and Piotr Bojanowski.
\newblock Dinov2: Learning robust visual features without supervision, 2024.

\bibitem{patashnik2021styleclip}
Or Patashnik, Zongze Wu, Eli Shechtman, Daniel Cohen-Or, and Dani Lischinski.
\newblock Styleclip: Text-driven manipulation of stylegan imagery.
\newblock In {\em Proceedings of the IEEE/CVF international conference on computer vision}, pages 2085--2094, 2021.

\bibitem{peng2024viteraser}
Dezhi Peng, Chongyu Liu, Yuliang Liu, and Lianwen Jin.
\newblock Viteraser: Harnessing the power of vision transformers for scene text removal with segmim pretraining.
\newblock In {\em Proceedings of the AAAI Conference on Artificial Intelligence}, volume~38, pages 4468--4477, 2024.

\bibitem{radford2021learning}
Alec Radford, Jong~Wook Kim, Chris Hallacy, Aditya Ramesh, Gabriel Goh, Sandhini Agarwal, Girish Sastry, Amanda Askell, Pamela Mishkin, Jack Clark, et~al.
\newblock Learning transferable visual models from natural language supervision.
\newblock In {\em International conference on machine learning}, pages 8748--8763. PMLR, 2021.

\bibitem{ramesh2021zero}
Aditya Ramesh, Mikhail Pavlov, Gabriel Goh, Scott Gray, Chelsea Voss, Alec Radford, Mark Chen, and Ilya Sutskever.
\newblock Zero-shot text-to-image generation.
\newblock In {\em International conference on machine learning}, pages 8821--8831. Pmlr, 2021.

\bibitem{rombach2022high}
Robin Rombach, Andreas Blattmann, Dominik Lorenz, Patrick Esser, and Bj{\"o}rn Ommer.
\newblock High-resolution image synthesis with latent diffusion models.
\newblock In {\em Proceedings of the IEEE/CVF conference on computer vision and pattern recognition}, pages 10684--10695, 2022.

\bibitem{Rombach_2022_CVPR}
Robin Rombach, Andreas Blattmann, Dominik Lorenz, Patrick Esser, and Bj\"orn Ommer.
\newblock High-resolution image synthesis with latent diffusion models.
\newblock In {\em Proceedings of the IEEE/CVF Conference on Computer Vision and Pattern Recognition (CVPR)}, pages 10684--10695, June 2022.

\bibitem{singh2021textocr}
Amanpreet Singh, Guan Pang, Mandy Toh, Jing Huang, Wojciech Galuba, and Tal Hassner.
\newblock Textocr: Towards large-scale end-to-end reasoning for arbitrary-shaped scene text.
\newblock In {\em Proceedings of the IEEE/CVF conference on computer vision and pattern recognition}, pages 8802--8812, 2021.

\bibitem{song2020denoising}
Jiaming Song, Chenlin Meng, and Stefano Ermon.
\newblock Denoising diffusion implicit models.
\newblock {\em arXiv preprint arXiv:2010.02502}, 2020.

\bibitem{Tang_2021}
Zhengmi Tang, Tomo Miyazaki, Yoshihiro Sugaya, and Shinichiro Omachi.
\newblock Stroke-based scene text erasing using synthetic data for training.
\newblock {\em IEEE Transactions on Image Processing}, 30:9306–9320, 2021.

\bibitem{tursun2020mtrnet++}
Osman Tursun, Simon Denman, Rui Zeng, Sabesan Sivapalan, Sridha Sridharan, and Clinton Fookes.
\newblock Mtrnet++: One-stage mask-based scene text eraser.
\newblock {\em Computer Vision and Image Understanding}, 201:103066, 2020.

\bibitem{tursun2019mtrnet}
Osman Tursun, Rui Zeng, Simon Denman, Sabesan Sivapalan, Sridha Sridharan, and Clinton Fookes.
\newblock Mtrnet: A generic scene text eraser.
\newblock In {\em 2019 International Conference on Document Analysis and Recognition (ICDAR)}, pages 39--44. IEEE, 2019.

\bibitem{10.5555/3295222.3295349}
Ashish Vaswani, Noam Shazeer, Niki Parmar, Jakob Uszkoreit, Llion Jones, Aidan~N. Gomez, \L{}ukasz Kaiser, and Illia Polosukhin.
\newblock Attention is all you need.
\newblock In {\em Proceedings of the 31st International Conference on Neural Information Processing Systems}, NIPS'17, page 6000–6010, Red Hook, NY, USA, 2017. Curran Associates Inc.

\bibitem{wang2024contextstablevisualconsistentimageinpainting}
Yikai Wang, Chenjie Cao, and Ke~Fan Xiangyang Xue~Yanwei Fu.
\newblock Towards context-stable and visual-consistent image inpainting, 2024.

\bibitem{wang2021pert}
Yuxin Wang, Hongtao Xie, Shancheng Fang, Yadong Qu, and Yongdong Zhang.
\newblock Pert: a progressively region-based network for scene text removal.
\newblock {\em arXiv preprint arXiv:2106.13029}, 2021.

\bibitem{wei2022masked}
Chen Wei, Haoqi Fan, Saining Xie, Chao-Yuan Wu, Alan Yuille, and Christoph Feichtenhofer.
\newblock Masked feature prediction for self-supervised visual pre-training.
\newblock In {\em Proceedings of the IEEE/CVF Conference on Computer Vision and Pattern Recognition}, pages 14668--14678, 2022.

\bibitem{wei2022mvp}
Longhui Wei, Lingxi Xie, Wengang Zhou, Houqiang Li, and Qi Tian.
\newblock Mvp: Multimodality-guided visual pre-training.
\newblock In {\em European conference on computer vision}, pages 337--353. Springer, 2022.

\bibitem{wimbauer2024cache}
Felix Wimbauer, Bichen Wu, Edgar Schoenfeld, Xiaoliang Dai, Ji Hou, Zijian He, Artsiom Sanakoyeu, Peizhao Zhang, Sam Tsai, Jonas Kohler, et~al.
\newblock Cache me if you can: Accelerating diffusion models through block caching.
\newblock In {\em Proceedings of the IEEE/CVF Conference on Computer Vision and Pattern Recognition}, pages 6211--6220, 2024.

\bibitem{yang2023paint}
Binxin Yang, Shuyang Gu, Bo Zhang, Ting Zhang, Xuejin Chen, Xiaoyan Sun, Dong Chen, and Fang Wen.
\newblock Paint by example: Exemplar-based image editing with diffusion models.
\newblock In {\em Proceedings of the IEEE/CVF Conference on Computer Vision and Pattern Recognition}, pages 18381--18391, 2023.

\bibitem{yang2020swaptext}
Qiangpeng Yang, Jun Huang, and Wei Lin.
\newblock Swaptext: Image based texts transfer in scenes.
\newblock In {\em Proceedings of the IEEE/CVF Conference on Computer Vision and Pattern Recognition}, pages 14700--14709, 2020.

\bibitem{strdd}
Wentao Yang, Hui Liu, and Ning Liu.
\newblock Strdd: Scene text removal with diffusion probabilistic models.
\newblock In Shuo Yang and Huimin Lu, editors, {\em Artificial Intelligence and Robotics}, pages 159--170, Singapore, 2022. Springer Nature Singapore.

\bibitem{yu2018generative}
Jiahui Yu, Zhe Lin, Jimei Yang, Xiaohui Shen, Xin Lu, and Thomas~S Huang.
\newblock Generative image inpainting with contextual attention.
\newblock In {\em Proceedings of the IEEE conference on computer vision and pattern recognition}, pages 5505--5514, 2018.

\bibitem{zdenek2020erasing}
Jan Zdenek and Hideki Nakayama.
\newblock Erasing scene text with weak supervision.
\newblock In {\em Proceedings of the IEEE/CVF Winter Conference on Applications of Computer Vision}, pages 2238--2246, 2020.

\bibitem{zhang2023adding}
Lvmin Zhang, Anyi Rao, and Maneesh Agrawala.
\newblock Adding conditional control to text-to-image diffusion models.
\newblock In {\em Proceedings of the IEEE/CVF International Conference on Computer Vision}, pages 3836--3847, 2023.

\bibitem{zhang2019ensnet}
Shuaitao Zhang, Yuliang Liu, Lianwen Jin, Yaoxiong Huang, and Songxuan Lai.
\newblock Ensnet: Ensconce text in the wild.
\newblock In {\em Proceedings of the AAAI conference on artificial intelligence}, volume~33, pages 801--808, 2019.

\bibitem{zhao2021large}
Shengyu Zhao, Jonathan Cui, Yilun Sheng, Yue Dong, Xiao Liang, Eric~I Chang, and Yan Xu.
\newblock Large scale image completion via co-modulated generative adversarial networks.
\newblock {\em arXiv preprint arXiv:2103.10428}, 2021.

\end{thebibliography}
}

\end{document}